\documentclass[conference]{IEEEtran}
\IEEEoverridecommandlockouts
\usepackage{cite}
\usepackage{amsmath,amssymb,amsfonts}
\usepackage{algorithmic}
\usepackage{graphicx}
\usepackage{hyperref}
\usepackage{multirow,textcomp}
\usepackage{xcolor}
\def\BibTeX{{\rm B\kern-.05em{\sc i\kern-.025em b}\kern-.08em
    T\kern-.1667em\lower.7ex\hbox{E}\kern-.125emX}}
\makeatletter 
\newcommand{\etal}{\textit{et al.}}
\newcommand{\linebreakand}{%
  \end{@IEEEauthorhalign}
  \hfill\mbox{}\par
  \mbox{}\hfill\begin{@IEEEauthorhalign}
}
\makeatother 

\begin{document}

\title{Data Quality Matters: Suicide Intention Detection on Social Media Posts Using RoBERTa-CNN}

\author{\IEEEauthorblockN{1\textsuperscript{st} Emily Lin\IEEEauthorrefmark{1}, 1\textsuperscript{st} Jian Sun\IEEEauthorrefmark{2}, 1\textsuperscript{st} Hsingyu Chen\IEEEauthorrefmark{2} and 2\textsuperscript{nd} Mohammad H. Mahoor\IEEEauthorrefmark{1},~\IEEEmembership{Senior Member,~IEEE}}
\IEEEauthorblockA{Email: Emily.lin@du.edu Jian.Sun86@du.edu Hsingyu.chen@du.edu mohammad.mahoor@du.edu}
\IEEEauthorblockA{\IEEEauthorrefmark{1}\textit{Department of Electrical and Computer Engineering, University of Denver}, Denver, United States of America}
\IEEEauthorblockA{\IEEEauthorrefmark{2}\textit{Department of Computer Science, University of Denver}, Denver, United States of America} 

\thanks{Emily, Jian, and Hsingyu are co-first authors with equal contribution.}
\thanks{Corresponding Author: Mohammad H. Mahoor. Tel: +1 (303) 871-3745. URL: \url{http://mohammadmahoor.com}}
\thanks{Jian Sun: ORCID: 0000-0002-9367-0892 URL: \url{https://sites.google.com/view/sunjian/home}}
}

\maketitle

\begin{abstract}
Suicide remains a pressing global health concern, necessitating innovative approaches for early detection and intervention. This paper focuses on identifying suicidal intentions in posts from the SuicideWatch subreddit by proposing a novel deep-learning approach that utilizes the state-of-the-art RoBERTa-CNN model. The robustly Optimized BERT Pretraining Approach (RoBERTa) excels at capturing textual nuances and forming semantic relationships within the text. The remaining Convolutional Neural Network (CNN) head enhances RoBERTa's capacity to discern critical patterns from extensive datasets. To evaluate RoBERTa-CNN, we conducted experiments on the Suicide and Depression Detection dataset, yielding promising results. For instance, RoBERTa-CNN achieves a mean accuracy of 98\% with a standard deviation (STD) of 0.0009. Additionally, we found that data quality significantly impacts the training of a robust model. To improve data quality, we removed noise from the text data while preserving its contextual content through either manually cleaning or utilizing the OpenAI API.
\end{abstract}


\section{Introduction} \label{sec:1}

Suicide is one of the leading causes of mortality worldwide with approximately 1.53 million cases in 2020~\cite{bertolote2002suicide}. The World Health Organization (WHO) estimated that approximately 703,000 people committed suicide worldwide in 2022~\cite{Suicide}. At least 1 out of 31 individuals with suicide ideation attempted to end their life~\cite {harmer2020suicidal}. Thus, suicide is an urgent public health concern worldwide, which requires early intervention~\cite{haque2020transformer}.  

Linguistic symptoms such as pessimistic note writing and remarks are common signs of committing suicide ideation~\cite{haque2020transformer}, which, thus, benefits suicide intention detection (SID) for early intervention. Nowadays, people are inclined to anonymously express themselves freely online instead of in the real world. Hence, it is more practical to research online content. At the same time, deep learning (DL) already paved the way for psychology research like SID~\cite{sawhney2020time}. Given the importance of SID and the power of DL, this paper presents our recent study on using natural language processing (NLP) models to research SID by analyzing linguistic symptoms from social media posts and comments (Reddit).  

Specifically, social media has an abundant linguistic corpus for NLP research~\cite{ji2020suicidal}. This paper utilized the Suicide and Depression Detection (SDD) dataset\footnote{\href{https://www..com/datasets/nikhileswarkomati/suicide-watch}{https://www..com/datasets/nikhileswarkomati/suicide-watch}}. The developer collected SDD by scrapping the posts and comments from the SuicideWatch and Depression subreddits of the Reddit Platform\footnote{\href{https://www.reddit.com/r/SuicideWatch/}{https://www.reddit.com/r/SuicideWatch/}}.

Currently, transformer-based NLP models are good at tackling various medical tasks (cardiology and neurology research~\cite{le2021machine}) and mental health tasks~\cite{zhang2022natural}. To capture the early signs of suicide ideation from linguistic data, we created a RoBERTa-CNN~\cite{haque2020transformer} model to conduct the study. RoBERTa-CNN, as a variant of BERT (Bidirectional Encoder Representations from Transformers), contains RoBERTa backbone and CNN head. The attention mechanism enables RoBERTa to comprehend complex contextual information and capture sophisticated linguistic patterns well~\cite{liu2019roberta}, while the CNN head strengthens the power to obtain linguistic patterns and extract speech-related features, making the model more robust and generalizable~\cite{song2019abstractive}. RoBERTa-CNN improves the performance and overcomes the constraints of the original RoBERTa model~\cite{liu2019roberta}. Specifically, RoBERTa eliminates the next sentence prediction and uses a larger dataset for the pre-trained model, which improves contextual awareness over typical models~\cite{liu2019roberta}. Compared to BERT and RoBERTa, RoBERTa-CNN handles long sequences more effectively~\cite{semary_improving_2023, liu2019roberta} due to its CNN head. Hence, RoBERTa-CNN is well-suited for SID.

RoBERTa-CNN achieved a high mean accuracy of 98\% in our experiments, which outperforms other text-based SID studies (see Table~\ref{tab:exp_reslt}). Furthermore, this solid performance demonstrates that RoBERTa-CNN predicts SID reliably as it is robust under different contexts and expressions.

This research also supports the significance of data quality. Our data cleaning process eliminates noisy information, thus enhancing data quality and contributing to superior performance. This underscores the essential role of providing high-quality data for training AI models to achieve high accuracy.

The main contributions of this paper are summarized below: 
\begin{itemize}
    \item We developed a RoBERTa-CNN model to do SID on the SDD dataset, an online corpus dataset.
    \item The proposed automatic SID system achieved a mean accuracy of 98\%, higher than that in the state-of-the-art methods on the same dataset. 
    \item We find the best text embedding length for the SDD dataset by fine-tuning the relative hyperparameter.
    \item We used OpenAI API to clean up the dataset due to the key of data quality in training RoBERTa-CNN.
\end{itemize}

The remainder of this paper is organized as follows. Section~\ref{sec:2} reviews the related works on SID. Section~\ref{sec:3} introduces the model. Section~\ref{sec:4} presents the experiments and the ablation study. Sections~\ref{sec:5} and~\ref{sec:6} cover the discussion and conclusion.

\section{Related Works} \label{sec:2}

Currently, BERT-based models~\cite{9392692} are good for SID~\cite{acheampong2021transformer}, which capture semantics and contextual features well~\cite{7384418}. RoBERTa, BERT's variant, applies dynamic masking to embed text instead of the static embedding used in the typical BERT. RoBERTa utilizes the large training batch size to avoid logic errors. Unlike BERT, RoBERTa reduces the warming-up steps and increases the learning rate~\cite{zhang2020accelerating,liu2019roberta}, which makes RoBERTa efficient~\cite{liu2019roberta}. 

RoBERTa succeeds in various NLP tasks such as text summarization~\cite{mishra2021looking,mohamed2023ssn} and text analysis~\cite{liao2021improved}. Specifically, Mishra~\etal~\cite{mishra2021looking} applied RoBERTa and achieved 86\% accuracy for the factual correctness of textual summarization. Mohamed~\etal~\cite{mohamed2023ssn} reported 76\% accuracy for text analysis by using RoBERTa.

Other BERT variants perform robustly on NLP tasks~\cite{9392692, martinez2020early, hassib2022aradepsu, 10037677} but have drawbacks. Compared with RoBERTa, ALBERT is time-consuming~\cite{sundelin2023comparing}, XLNet is slower too~\cite{liu2019roberta}, and DistillBERT loses information due to the reduction of model size~\cite{9392692}.

In addition, much work has focused on analyzing keywords instead of the entire sentence~\cite{zhang2021automatic}, which causes information loss. Hence, this study uses RoBERTa-Base as the backbone to do sentence analysis.

\section{Methodology} \label{sec:3}

This paper applies RoBERTa-CNN, which involves a RoBERTa backbone and a CNN head. RoBERTa extracts textual features from the dataset (Sections~\ref{sec:3.1} and~\ref{sec:3.2}). The CNN classifier processes the extracted features and enhances the prediction performance (Section~\ref{sec:3.3}).

\begin{figure*}[t]
\centering
\resizebox{17cm}{!}{
\vspace{-0.8cm}
\includegraphics[width=0.7\textwidth]{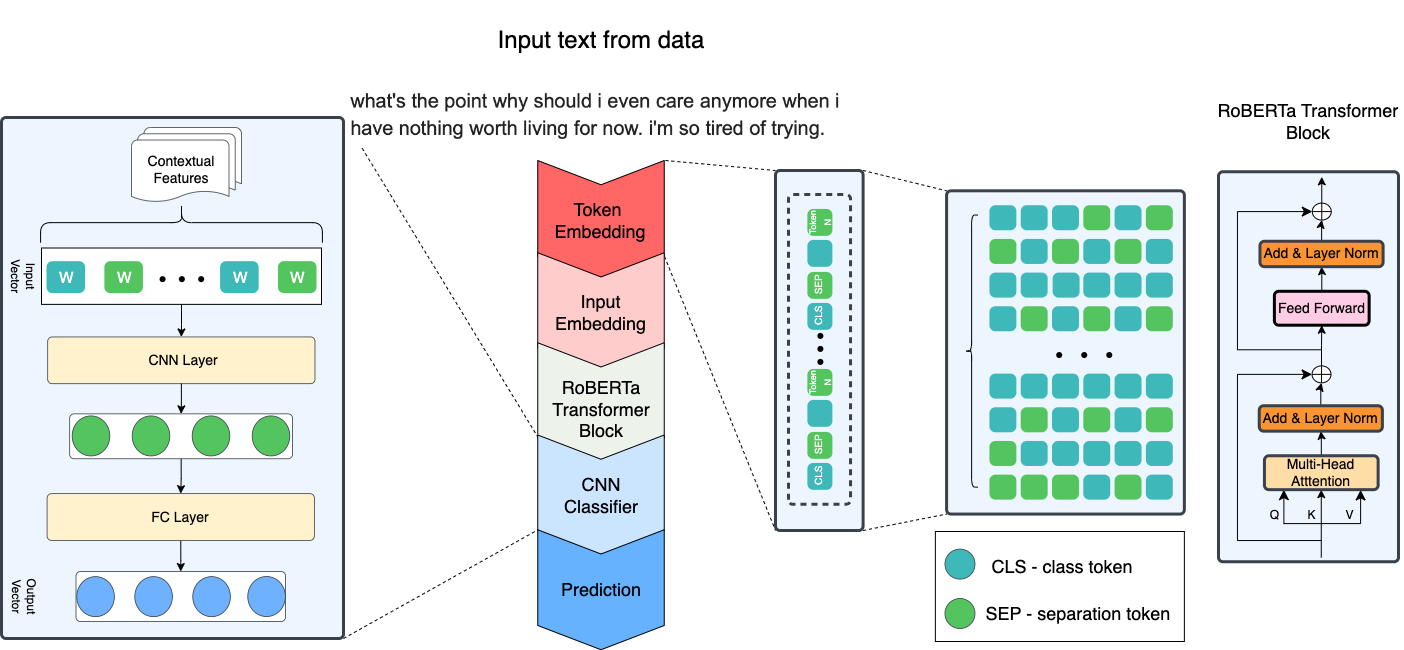}}
\vspace{-0.2cm}
\caption{shows the structure of RoBERTa-CNN, which contains the RoBERTa backbone and CNN head.}
\label{fig:model_structure}
\vspace{-0.5cm}
\end{figure*}

\subsection{RoBERTa Embedding} \label{sec:3.1}

The RoBERTa embedding tokenizes input comments (individuals' posts on Reddit), $X=[x_{1},x_{2}, \dotsm, x_{n}]$, where $n$ is the number of sentences, $x_{i}$ has the size of [Max\_Len], $i\in [1,n]$, Max\_Len is the token length of $x_{i}$. The shape of $X$ is [n, Max\_Len].

The RoBERTa embedding is a context-dependent word embedding~\cite{liu2019roberta}, which considers word order within the context and learns context-specific nuances~\cite{haque2020transformer, liu2019roberta}. Thereby, $X$ contains many dynamic and robust representations. Moreover, the RoBERTa embedding is pre-trained on large text corpora.

\subsection{RoBERTa Backbone} \label{sec:3.2}

Fig.~\ref{fig:model_structure} shows that RoBERTa passes the input $X$ into the transformer encoder where the core is the attention mechanism~\eqref{eq:attn}. Given a layer $i$, the computing processes are~\eqref{eq:layer1} and~\eqref{eq:layer2}.
\vspace{-0.2cm}
\begin{align}
\text{Attention}(Q,K,V) &= \text{Softmax}(\frac{Q\cdot K^{T}}{\sqrt{d_{K}}})V \label{eq:attn}\\
\text{Attn}_{out} &= LN(MHSA(L_{i-1})+L_{i-1}) \label{eq:layer1}\\
L_{i} &= LN(FF(\text{Attn}_{out})+\text{Attn}_{out}) \label{eq:layer2}
\end{align}
where $L_{i-1}$ is the output of layer $i-1$ and the input of this layer $i$, MHA is multi-head self-attention module, LN represents layer normalization, $\text{Attn}_{out}$ is the output of the attention module. FF is the feed-forward module, and $L_{i}$ is the output of layer $i$. Its shape is [n, Max\_Len, 768], where 768 is the hidden dimension within the attention mechanism.

While analyzing text data, the attention mechanism focuses more on suicidal intention-related words, which benefits RoBERTa-CNN in identifying imperceptible but valuable linguistic signs well ~\cite{haque2020transformer,liu2019roberta}.

\subsection{CNN head} \label{sec:3.3}

The CNN head contains one Conv layer, one ReLU layer, one 1D-max pooling (Max\_Pool1D) layer, and one fully-connected (FC) layer (See Fig.~\ref{fig:model_structure}), which is summarized as~\eqref{eq:cnn_part} and~\eqref{eq:y_hat}. The CNN head's input is RoBERTa's output $L_{N}$ after swapping the last two dimensions. $N$ is the number of attention layers. The size of $L_{N}$ is [n, 768, Max\_Len]. The CNN head outputs $\hat{y}$ with the shape of [n, 2].
\vspace{-0.1cm}
\begin{align}
\text{CNN\_Out} &= \text{Max\_Pool1D}(\text{ReLU}(\text{Conv}(L_{N}))) \label{eq:cnn_part}\\
\hat{y} &= \text{FC}(\text{Squeeze}(\text{CNN\_Out})) \label{eq:y_hat}
\end{align}

The kernel size and channel number of the Conv layer are 2 and 100. The size of CNN\_Out is [n, 100, 1]. Squeeze function reduces CNN\_Out from a 3D to a 2D tensor with the size of [n, 100]. Then, the FC layer deducts dimensions from 100 to 2. The Conv layer enables RoBERTa-CNN to pay attention to the local features and patterns that are key for the classification.

\section{Experiment} \label{sec:4}
This section first introduces the SDD dataset and the experiment settings. Then, it presents and analyzes the experimental results and ablation study.

\subsection{Dataset} \label{sec:4.1}

The Suicide and Depression Detection (SDD) dataset consists of the online posts and comments from the SuicideWatch and Depression channel, subreddits of the Reddit Platform. The time range is from December 2018 to January 2021. Table~\ref{tab:data_sample} shows some data samples. We randomly picked 110,040 out of 232,074 posts with two categories (Suicide and non-Suicide).

\begin{table} [t]
\centering
\caption{Data Samples of the Suicide and Depression
Detection Dataset}
\vspace{-0.2cm}
\resizebox{8.5cm}{!}{
\begin{tabular}{l|l}
\hline
\multirow{6}{1.5cm}{Suicide} & \multirow{2}{7cm}{1. I feel worthless and useless to everyone. I'm a burden what's the point of being here?} \\
 & \\
 & \multirow{2}{7cm}{2. I feel like overdosing again. I can't live like this.} \\
 & \\
 & 3. I feel bad. I just wanna stop existing. \\
\hline
\multirow{8}{1.5cm}{Non-suicide} & \multirow{3}{7cm}{1. Would you ever trust your school therapist? I myself wouldn't tbh, there doesn't seem to be a lot of confidentiality.} \\
 & \\ & \\
 & \multirow{2}{7cm}{2. Is it just me or is apple cider really good? it is one of the best drinks I think.} \\
 & \\
 & \multirow{3}{7cm}{3. Why is everyone talking about us? I'm genuinely confused what exactly did we do?} \\
 & \\ & \\
\hline
\end{tabular}}\label{tab:data_sample} 
\vspace{-0.5cm}
\end{table}

\textbf{Data processing}: Three data cleaning steps are: 1) remove all noise symbols (accents, hyperlinks, and emoticons); 2) remove all non-English characters; and 3) delete all sentence fragments. Since the research involves sentence analysis, normal standard preprocessing was avoided, like converting all text to lowercase, removing punctuation and accents, stripping white space, and stop-words. We manually cleaned 50,000 samples and applied OpenAI API to clean 60,040 others. There are 35,270 suicide samples and 74,770 non-suicide ones.

\begin{table} [t]
\centering
\caption{The subset division of Suicide and Depression Detection.}
\vspace{-0.2cm}
\resizebox{8.5cm}{!}{
\begin{tabular}{lllll}
\hline
 Dataset & Overall & Train & Validation & Test  \\
\hline
\multirow{2}{2.7cm}{Suicide and Depression Detection} & \multirow{2}{1.cm}{110,040} & \multirow{2}{1.cm}{88,032} & \multirow{2}{1.cm}{11,004} & \multirow{2}{1.cm}{11,004} \\
&&&& \\
\hline
\end{tabular}}\label{tab:exp_data}
\vspace{-0.5cm}
\end{table}

Then, following the pattern of conductive learning, we randomly and stratifiedly split the dataset into non-overlapped subsets, train, validation, and test sets, with the ratio of 8:1:1~\cite{9660152}. Table~\ref{tab:exp_data} presents the distribution.

\subsection{Implementation detail} \label{sec:4.2}

For training, we first augmented the text with normalization, retrieval of synonyms, and synonym replacement. The batch size was 200. We initialized the learning rate to 1e-6 for the Adam optimizer. Conductive learning tests if RoEBRTa-CNN can detect unseen data well. Binary Cross-entropy loss controls the training. Then, we randomly trained RoBERTa-CNN (over 35 epochs) on the SID dataset using an NVIDIA RTX 3090 GPU five times and reported the mean and standard deviation (STD) values.

\subsection{Evaluation metrics} \label{sec:4.3}

The used metrics are Accuracy, Precision, Recall, F1 Score, and AUC (Area Under the Receiver Operating Characteristic Curve). Higher metric values mean better RoBERTa-CNN.

\subsection{Experiment by RoBERTa-CNN} \label{sec:4.4}

\begin{table}[t]
\centering
\caption{The experimental results on SDD dataset.}
\vspace{-0.2cm}
\resizebox{9cm}{!}{
\begin{tabular}{llllll}
\hline
 Models & Accuracy(\%) & Precision(\%) & Recall(\%) & F1 Score(\%) & AUC (\%) \\
\hline
\multirow{2}{1.3cm}{CNN-BiLSTM~\cite{AldhyaniandAlsubari(2022)}} & \multirow{2}{1.cm}{95.00} & \multirow{2}{1.cm}{94.30} & \multirow{2}{1.cm}{94.90} & \multirow{2}{1.cm}{95.00} & \multirow{2}{1cm}{-} \\
&&&&& \\
\multirow{2}{1.3cm}{XGBoost~\cite{AldhyaniandAlsubari(2022)}} & \multirow{2}{1.cm}{91.50} & \multirow{2}{1.cm}{93.50} & \multirow{2}{1.cm}{89.10} & \multirow{2}{1.cm}{91.30} & \multirow{2}{1.cm}{-} \\
&&&&& \\
\multirow{3}{1.3cm}{EFS + pBGSK + SVM~\cite{priya2023contemporarymulti}} & \multirow{3}{1.cm}{96.20} & \multirow{3}{1.cm}{-} & \multirow{3}{1.cm}{-} & \multirow{3}{1.cm}{92.90} & \multirow{3}{1.cm}{-} \\
&&&&& \\
&&&&& \\
\hline
Ours & \textbf{98.00$\pm$0.09} & \textbf{96.98$\pm$0.22} & \textbf{96.64$\pm$0.30} & \textbf{96.81$\pm$0.14} & \textbf{97.63$\pm$0.13} \\
\hline
\end{tabular}}\label{tab:exp_reslt}
\vspace{-0.5cm}
\end{table}

The SDD dataset is to evaluate RoBERTa-CNN on SID. Table~\ref{tab:exp_reslt} implies that RoBERTa-CNN achieved 98.00\% mean accuracy with an STD of 0.09\%, which is 3\%, 6.5\%, and 1.8\% higher than the accuracy on CNN-BiLSTM, XGBoost, and EFS + pBGSK + SVM, respectively. Then, the overall highest Precision, Recall, F1 Score, and AUC also indicate that RoBERTa-CNN is very accurate and robust at detecting both suicide and non-suicide intention. In addition, the conductive learning setting makes the experimental result more objective and convincing. Hence, RoBERTa-CNN predicts SID well and surpasses CNN-BiLSTM, XGBoost, and EFS + pBGSK + SVM on the SDD dataset.

\subsection{Ablation study} \label{sec:4.5}

\subsubsection{The Effectiveness of Max\_Len} \label{sec:4.5.1}

Each sample's token length $Max\_Len$ significantly affects the model performance. By setting Max\_Len as $[64, 128, 256]$, we did an ablation study to find optimized Max\_Len. 

\begin{table} [t]
\centering
\caption{The ablation study results. The upper table is to tune Max\_Len, and the bottom one is to compare with the origin RoBERTa. The value format is Mean$\pm$STD.}
\vspace{-0.2cm}
\resizebox{9cm}{!}{
\begin{tabular}{llllll}
\hline
Max\_Len & Accuracy (\%) & Precision(\%) & Recall (\%) & F1 Score (\%) & AUC(\%) \\
\hline
64 & 97.93$\pm$0.10 & \textbf{96.99$\pm$0.45} & 96.46$\pm$0.63 & 96.72$\pm$0.15 & 97.54$\pm$0.22 \\
128 & 97.98$\pm$0.06 & 96.74$\pm$0.41 & \textbf{96.87$\pm$0.28} & 96.80$\pm$0.09 & \textbf{97.68$\pm$0.06} \\
256 & \textbf{98.00$\pm$0.09} & 96.98$\pm$0.22 & 96.64$\pm$0.30 & \textbf{96.81$\pm$0.14} & 97.63$\pm$0.13 \\
\hline
Classifier & Accuracy (\%) & Precision(\%) & Recall (\%) & F1 Score (\%) & AUC(\%) \\
\hline
RoBERTa & 97.95$\pm$0.06 & \textbf{97.03$\pm$0.39} & 96.70$\pm$0.39 & \textbf{96.86$\pm$0.05} & 97.66$\pm$0.11 \\
RoBERTa-CNN & \textbf{97.98$\pm$0.06} & 96.74$\pm$0.41 & \textbf{96.87$\pm$0.28} & 96.80$\pm$0.09 & \textbf{97.68$\pm$0.06} \\
\hline
\end{tabular}\label{tab:ablation}}
\vspace{-0.5cm}
\end{table}

Table~\ref{tab:ablation} reveals that setting Max\_Len = 256 leads to the best performance with a Top-1 accuracy of (98.00$\pm$0.09)\% and a Top-1 F1 Score of (96.81$\pm$0.14)\%. Furthermore, setting Max\_Len = 256 achieves comparable Precision, Recall, and AUC, which are very close to the highest values. For example, the mean Precision (96.98\%) when Max\_Len = 256 is merely 0.01\% lower than that of Max\_Len = 64, and the STD of Max\_Len = 256 (0.22\%) is 0.23\% lower than that of Max\_Len = 64 (0.45\%). Hence, Max\_Len = 256 contributes to a more robust Precision value among five experiments than Max\_Len = 64. In general, choosing Max\_Len = 256 suits the model performance more.

\subsubsection{The Effectiveness of CNN Classifier} \label{sec:4.5.2}

This experiment tests whether or not the CNN classifier strengthens RoBERTa. To speed up the experiment, we set Max\_Len to 128.

Table~\ref{tab:ablation} implies that RoBERTa-CNN (97.98\%) reaches better mean accuracy than RoBERTa (97.95\%). RoBERTa-CNN's mean AUC and mean Recall are higher than RoBERTa's, which indicates that the CNN classifier detects the positive and negative samples better than the classifier used by RoBERTa. Moreover, RoBERTa-CNN's solid Precision and F1 Score are comparable to RoBERTa's. Hence, RoBERTa-CNN, in comparison, is better than RoBERTa on the SDD dataset. 

Overall good results of the ablation study show that RoBERTa-CNN is robust on the SDD dataset.

\section{Discussion} \label{sec:5}

This paper uses RoBERTa-CNN to conduct SID because of its effectiveness in language tasks. The results in Section~\ref{sec:4.4} show its power on text data. The RoBERTa model with its attention mechanism is powerful in capturing contextual information. CNN reduces the limitation of the RoBERTa model by capturing the local features with its Conv layer and identifies insight linguistic patterns in a sentence~\cite{wang2020short}. Thus, the CNN head enhances the robustness and accuracy of RoBERTa-CNN. 

Moreover, the data cleaning removes the noise from the original SDD dataset and improves the data quality, substantially contributing to RoBERTa-CNN’s solid performance.

Nevertheless, RoBERTa-CNN has some limitations. Due to its complexity, RoBERTa-CNN requires huge computational costs as the Max\_Len increases. Light-weight RoBERTa may solve the problem. The simple structure of the CNN head drives us to find more power classifiers such as XnODR/XnIDR~\cite{sun2023xnodr} and Multi-channel classifier~\cite{sun2024mc}. Finally, there is very little ChatGPT-related research on SID now, which motivates us to use the ChatGPT model to do contrastive experiments on the SID task.
 
\section{Conclusions} \label{sec:6}

This paper presented a study on SID leveraging RoBERTa-CNN. We tested the model on the SDD dataset. The promising results imply the robustness of the model in identifying the textual signs of suicide ideation, convincing us that RoBERTa-CNN does SID well on text data. Future work includes developing a multi-modal model to do SID by analyzing visual, text, and speech data.

\section*{Acknowledgment}

The author would like to thank Miss. Zhuo Chen for helping clean the data, and Mr. Jiahan Li for joining weekly meetings and contributing to the discussions. The research was partially supported by a Partners in Scholarship (PINs) summer research support to Emily Lin.

\bibliographystyle{unsrt} 
\bibliography{ref}

\end{document}